\documentclass{article}

\usepackage{multirow}
\usepackage[utf8]{inputenc}
\usepackage[small]{dgruyter}
\usepackage{microtype}
\usepackage[autostyle=true]{csquotes}
\usepackage{graphicx}
\usepackage{todonotes}

\usepackage{soul}
\usepackage{xcolor}

\usepackage[backend=bibtex, style=authoryear]{biblatex}
\begin{document}

  \articletype{Paper}
  \author[1]{Lucas Rafael Stefanel Gris}
  \author*[2]{Ricardo Marcacini}
  \author[3]{Arnaldo Candido Junior} 
  \author[4]{Edresson Casanova}
  \author[1]{Anderson Soares}
  \author[2]{Sandra Maria Aluísio}
  \runningauthor{Gris et al.}
  \affil[1]{Federal University of Goiás, Brazil}
  \affil[2]{University of São Paulo, Brazil}
  \affil[3]{São Paulo State University, Brazil}
  \affil[4]{Coqui.ai, Germany}
\title{Evaluating OpenAI's Whisper ASR for Punctuation Prediction and Topic Modeling of life histories of the Museum of the Person}
  \runningtitle{Evaluating OpenAI's Whisper ASR}

\abstract{Automatic speech recognition (ASR) systems play a key role in applications involving human-machine interactions. Despite their importance, ASR models for the Portuguese language proposed in the last decade have limitations in relation to the correct identification of punctuation marks in automatic transcriptions, which hinder the use of transcriptions by other systems, models, and even by humans. However, recently Whisper ASR was proposed by OpenAI, a general-purpose speech recognition model that has generated great expectations in dealing with such limitations. This chapter presents the first study on the performance of Whisper for punctuation prediction in the Portuguese language. We present an experimental evaluation considering both theoretical aspects involving pausing points (comma) and complete ideas (exclamation, question, and fullstop), as well as practical aspects involving transcript-based topic modeling --- an application dependent on punctuation marks for promising performance. We analyzed experimental results from videos of Museum of the Person, a virtual museum that aims to tell and preserve people’s life histories, thus discussing the pros and cons of Whisper in a real-world scenario. Although our experiments indicate that Whisper achieves state-of-the-art results, we conclude that some punctuation marks require improvements, such as exclamation, semicolon and colon.}
\maketitle

\section{Introduction} 
\label{sec:sec1}
The conventional output of Automatic Speech Recognition (ASR) systems, both commercial and open-source, is a sequence of words without punctuation and capitalization, also called a raw output.
Examples of commercial ASR systems are Google Cloud Speech-to-Text\footnote{cloud.google.com/speech-to-text}, Microsoft Azure Speech Services\footnote{azure.microsoft.com/en-us/products/cognitive-services/speech-services/}, IBM Watson Speech to Text\footnote{www.ibm.com/br-pt/cloud/watson-speech-to-text}, and SpeechMatics\footnote{www.speechmatics.com/}. 
As for open-source ones we can cite  the Kaldi project \parencite{Povey_ASRU2011}, Mozilla's DeepSearch \parencite{Hannun2014},  NVIDIA's OpenSeq2Seq \parencite{openseq2seq2018}, and Facebook AI's  Wav2Letter++ \parencite{Pratap_2019_ICASSP}, wav2vec 2.0 \parencite{Baevski_2021_CORR} and wav2vec Unsupervised  \parencite{Baevski2021_neurips} models. 
Although this type of output is sufficient for several applications, such as voice commands, virtual assistants, and subtitles transcription where speech segments are usually short and/or independent, it is not adequate for applications that transcribe long segments of spontaneous or read speech. 

Differently from the previous systems cited,  Amazon Transcribe\footnote{docs.aws.amazon.com/transcribe/} tries to improve the understanding of the output of its ASR for English and German languages, by adding capitalization, punctuation and number formatting. In their site\footnote{docs.aws.amazon.com/transcribe/latest/dg/how-numbers.html} there are the following transcriptions for the spoken excerpt  ``Meet me at eight-thirty AM on June first at one-hundred Main Street with three-dollars-and-fifty-cents and one-point-five chocolate bars'':

\vspace{0.5cm}
\begin{description}
    \item[English and German transcription] Meet me at 8:30 a.m. on June 1st at 100 Main Street with \$3.50 and 1.5 chocolate bars
    \item[Other supported languages] Meet me at eight thirty a m on June first at one hundred Main Street with three dollars and fifty cents and one point five chocolate bars
\end{description}

Note that numbers were transcribed into their word forms for the languages supported by Amazon Transcribe but for English and German, where there are special number formatting decisions according to their meaning: time, data and cardinal numbers greater than ten are converted from word forms to Arabic numerals, monetary words are converted to symbols and  ``point'' (or ``dot'') are displayed as decimal numbers, respectively.

While number formatting enhances the understanding of a transcription,  ASR results are usually input for machine translation and name entity recognition (NER) systems that need sentence punctuation and word capitalization to help to improve their performance or to provide a correct output at all. For example, for the NER task the main feature used to find and classify entities into categories such as person names, organizations and locations is word capitalization. Also,  the transcript-based video topic modeling task is very useful for organizing, summarizing, and visualizing video content across similar subjects. In this task, sentence punctuation plays a key role to extract coherent segments from a video and generate video snippets of a given topic without truncation. 

The TaRSila project\footnote{sites.google.com/view/tarsila-c4ai} aims at growing speech datasets for Brazilian Portuguese language, looking to achieve state-of-the-art results for the task of ASR and also foresees applications in automatic organization of speech datasets into topics, to cite a few examples of studies. 
Datasets used to train ASRs are composed of speech segments aligned with their transcriptions. Currently, two large corpora of spontaneous speech are being processed: NURC-SP corpus\footnote{https://nurc.fflch.usp.br/} \parencite{CM2022, Gris_etal_2022} and 300-hour of life histories of the Museum of the Person (MuPe)\footnote{https://museudapessoa.org/}, a project launched on December 2022. 

As information in spoken language is transmitted through words amalgamated  with several acoustic cues, such as pitch, volume, speech rate, rhythm and timbre, phrases bounded by prosodic cues can convey coherent messages with a variety of linguistic functions, including the sentence mode: imperative, interrogative, assertive or exclamatory. These phrases are often called as intonational phrases or intonation units  \parencite{Biron2021}. According to \textcite{Biron2021}, the lengthening of speech rate at the end of an unit together with acceleration at its beginning, denoted as discontinuities in speech rate,  is a salient signal for identifying boundaries. 
Using two acoustic cues related to timing (discontinuities in  speaking rate and silent pauses), they proposed a heuristic method, using the output of an ASR system,  to identify boundaries in spontaneous speech. The heuristic boundary detection method of \textcite{Biron2021} is being adapted for Brazilian Portuguese (BP) to annotate terminal and non-terminal prosodic boundaries described in \textcite{Mello_etal_2012}, using the phonetic aligner UFPAlign \parencite{Batista22a}. The objective of this adaptation is to automatically segment the NURC-SP corpus to facilitate manual revision of its prosodic boundaries. 

It is expected that an effective automatic identification of prosodic boundaries will facilitate linguistic studies on spontaneous speech, help to create more useful datasets to train ASR models and extend the power of speech-related applications working on spontaneous speech. However, there are applications that also require a proposal of punctuation and truecasing for spontaneous speech. This is the case of the project that deals with life histories of MuPe.

MuPe is a virtual museum that aims to tell and preserve people's life histories and encourages the participation of people of different ages, genders, races and professions. The life histories are video (or audio) interviews with the prevalence of spontaneous speech. Founded in 1991, MuPe currently contains a rich and extensive digital collection of interviews with spontaneous speech in Portuguese that is potentially useful for various natural language processing tasks, especially automatic speech recognition.
For this project, a solution that would improve readability for people reading life histories and other systems that will process speech transcriptions would be: (1) split the long segments into sentences --- a task called sentence boundary detection --- and up-case the first word of each sentence; (2) up-case person's names and give the correct capitalization for named entities in general.

In this chapter, we will focus on the first experiments towards attending the needs of this new project. Our objective is to evaluate the use of the recently trained open-source ASR system called Whisper\footnote{https://openai.com/blog/whisper/} \parencite{radford2022robust} for both: (i) generating more easily readable transcripts for MuPe life histories and (ii) improving the transcript-based video topic modeling task on MuPe life histories. We compiled a gender balanced test dataset with life histories samples of men and women to: (i) evaluate  possible  gender differences in the use of question and exclamation marks, mainly; (ii) verify which punctuation marks generated by Whisper are best predicted in MuPe dataset; (iii) foresee possible improvements for this ASR to deal with spontaneous speech in Brazilian Portuguese.  In addition to assessing Whisper's ability to correctly identify punctuation in Portuguese transcription, we also investigated Whisper ASR in an application involving transcript-based video topic modeling. In this case, punctuation is used to identify coherent segments in the interview, i.e., segments with well-defined boundaries. Each segment is associated with a topic, and the most representative segments are used to obtain a summarized video containing the main themes discussed in the interview involving MuPe's life stories. In this way, we evaluate theoretical and practical aspects of the Whisper ASR punctuation marks. This is the first work, to the best of our knowledge, to evaluate Whisper for videos and audios in Brazilian Portuguese. 

Whisper was trained by the Artificial Intelligence research company OpenAI using a large multilingual dataset collected from the web.  Whisper has a new architecture trained to predict several tasks: (i) voice activity detection, that instructs the model to work only when a specific human language is playing, being robust when dealing with background noise/music; (ii) multilingual speech transcription with punctuation and (iii) speech translation to English. Whisper is a strong candidate ASR to deal with Portuguese speech because about a third of its audio training dataset is non-English, and Portuguese is the sixth language with more data for Multilingual Speech Recognition. Whisper was chosen in this study to evaluate the three tasks of the MuPe project: automatic transcription of spontaneous speech, speech segmentation with punctuation and transcript-based video topic modeling task. As a matter of fact, when using Whisper, automatic transcription and speech segmentation with punctuation is reduced to a sole task: automatic transcription with punctuation.

Examples of transcript-based video topic modeling are available on this project's page\footnote{https://googledrive}. The evaluation dataset and source code are publicly available at https://github.com/nilc-nlp/ for the reproducibility of the experimental evaluation.

In the following sections, we present related work on punctuation prediction mainly for spontaneous speech (Section \ref{sec:sec21}), and on transcript-based video topic modeling (Section \ref{sec:sec22}). In Section \ref{sec:sec3} we detail the features of the newly available ASR Whisper, contrasting the solution it proposes with previous ASR models  (Section \ref{sec:sec31}) and also run Whisper in an excerpt of a MuPe life history (Section \ref{sec:sec32}). Section \ref{sec:sec4} presents the experimental setup, including the evaluation dataset (Section \ref{sec:sec41}), data preparation and evaluation metrics for both tasks evaluated: punctuation prediction (Section \ref{sec:sec421}) and transcript-based video topic modeling (Section \ref{sec:sec422}) in the MuPe project. Section \ref{sec:sec5} brings results and discussions of the experiments. Section \ref{sec:sec6} concludes the chapter and presents future work we intend to pursue in this project.

\section{Related Work} 
\label{sec:sec2}

\subsection{Punctuation Prediction: Approaches, Datasets, Features, Evaluation Metrics and Results} 
\label{sec:sec21}

The output of conventional ASR systems is one of the main sources of data requiring capitalization and punctuation. When the output is a written text read aloud (read speech), the task is called original punctuation restoration, and for conversational/spontaneous speech the task — our interest in this chapter — is called punctuation prediction. In any case, not only ASR systems require capitalization and punctuation to facilitate understanding but there are also other sources of data in equal need: text obtained via optical character recognition (OCR), short text messages (SMS), tweets and the output text of conventional chatbots \parencite{survey_2022}.

The mainstream approach in the literature, known as a cascade approach,  is to  train a speech recognition model and a punctuation prediction model separately,
and then cascade them together, i.e. to insert punctuation marks into the transcription generated by the ASR as a post-processing step. Regarding features, three kinds of features used to predict/restore punctuation marks are lexical, acoustic, and the combination of acoustic and lexical features. In this section, we detail five recent works that use lexical features or multimodal frameworks for identifying punctuation in order to propose the evaluation of Whisper in the MuPe project and also to compare and contrast our results with the literature on punctuation restoration/prediction.

Recent lexical approaches in the literature for punctuation restoration use deep neural networks. The approaches vary from the use of pre-trained word embeddings,  attention mechanism, transformer based approaches trained on large text corpora — using only the pre-trained BERT model \parencite{DevlinCLT19} or pursuing an evaluation of different transformer based models.  

\textcite{alam-etal-2020-punctuation} explored different architectures and fine-tuned pre-trained models for punctuation restoration focusing on two languages: a high-resource (English) and a low-resource (Bangla). They evaluated monolingual language models (BERT, RoBERTa, ALBERT, DistilBERT) for English and multi-lingual language models (mBERT, XLM-RoBERTa) available in the HuggingFace’s Transformers library \parencite{wolf-etal-2020-transformers}. 
They also proposed a data augmentation strategy to cope with the problem of training the punctuation restoration model on clean texts but using it on noisy ASR texts. In this scenario the performance of the punctuation restoration model  may degrade due to errors of ASR models during recognition (insertion, substitution, and deletion of words). Their data augmentation strategy simulates these errors and dynamically creates a new sequence on the fly in a batch.
Their  setup for evaluation uses precision (P), recall (R), F1-score (F1) to evaluate the performance of the four labels: Comma, Period, Question and O (no punctuation mark followed), in the test sets of the ASR dataset \parencite{che_etal_2016_punctuation} from the International Workshop on Spoken Language Translation (IWSLT 2011). Therefore, it was possible to compare their results with the other five works competing on IWSLT 2011. For English, they obtained overall best results on the Reference and ASR transcriptions using the augmentation technique coupled with the RoBERTa-large model (82.9\% overall F1 on Reference test set and 74\% overall F1 on ASR test set). Results on Reference test set for Comma were the best: 76.7\% F1; for Period, F1 was 88.9\% and for Question 87.8\%. Results on ASR test set for Period were the best (82.3\%); for Comma and Question it was obtained 66.3\% and 63.4\%, respectively.
Analyzing the confusion matrix (in percentage), for ASR transcriptions, they found a high proportion of cases were Question and Comma are predicted as O and Period. 

However, using only lexical features makes the system vulnerable to speech recognition errors.  One possible improvement regarding the kind of features is to evaluate the combination of acoustic (prosodic) and lexical features to help surpass the problem of pauses in unnatural places in real ASR systems if only prosodic features are used. 

\textcite{Yi2019SelfattentionBM} comments that using lexical data with the corresponding speech data for training is also problematic as aligned data are limited for some languages. 
In this scenario, they proposed a self-attention based model using both word and speech embeddings, respectively  Glove \parencite{pennington_etal_2014_glove} and Speech2Vec \parencite{chung18c_interspeech}, solving the problem of dependence of speech data aligned with its transcription.  
Their self-attention based model can use any kind of textual and speech data.
They evaluated four self-attention based model combination trained to predict punctuation marks in their experiments: (1) Self-attention: the model is trained using the input embeddings learned in \textcite{NIPS2017}; (2) Self-attention-word: the model is trained using the word embeddings obtained from the pre-trained Glove embeddings; (3) Self-attention-speech: the model is trained using the speech embeddings obtained from the pre-trained Speech2Vec model; and (4) Self-attention-word-speech: the model is trained using the word embeddings and speech embeddings obtained from the pre-trained Glove and Speech2Vec, respectively; the word and speech embeddings are summed. 

All the models were evaluated using precision (P), recall (R), F1-score (F1). They evaluated performance for four labels: (i) Comma: includes commas, colons and dashes, (ii) Period: includes full stops, exclamation marks and semicolons, (iii) Question: only question mark, and (iv) O: no punctuation mark followed.
Like \textcite{alam-etal-2020-punctuation}, they also conducted experiments on the English  IWSLT 2011  ASR dataset (Reference and ASR)\footnote{http://hltc.cs.ust.hk/iwslt/index.php/evaluation-campaign/ted-task.html}.
Their overall results for the best performance model (self-attention-word-speech) on Reference and ASR are 72.9\% and 68.8\% F1, respectively. Results for each punctuation mark on Reference are: 64.1\% F1 for comma, 79.9\% for period, and 74.8\% for question. Results on ASR are:  61.7\% F1 for comma, 75.6\% for period, and 69.1\% for question.
The self-attention based model trained using word and speech embedding features outperforms the other three self-attention models they trained, but the results of \textcite{alam-etal-2020-punctuation} are still better than \textcite{Yi2019SelfattentionBM} (except for question mark),  probably due to the fact that \textcite{alam-etal-2020-punctuation} uses an augmentation technique, which improves performance on noisy ASR texts, and a transformer based model.

While  \textcite{Yi2019SelfattentionBM} conducted their evaluation on English IWSLT dataset which contain TED talks (a text genre related to prepared speech), the following two works \parencite{ZelaskoSMSCD18,sunkara20_interspeech} are more closely related to our interest — to evaluate punctuation prediction on conversational/spontaneous speech. 

\textcite{ZelaskoSMSCD18} reinforces that the problem of the task of punctuation prediction for conversational speech is the lack of reference datasets. While large corpus of written language available on the Web can be used to train a model, these are not representative of the conversational language and to annotate speech transcripts with proper punctuation is a time-consuming task.
They evaluated two deep neural networks models: one based on Convolutional Neural Nets (CNN) and the other based on Bidirectional Long Short-Term Memory (Bi-LSTM) networks where the input layer is a concatenation of two types of features: word embeddings to make the models more robust to different conversation topics; and a conversation side indicator, a kind of prosodic features related to timing.
The word embeddings are 300-dimensional pre-trained GloVe \parencite{pennington_etal_2014_glove} embeddings, trained on Common Web Crawl data. They also trained GloVe embeddings on conversational-like data (around 525M words) gathered by the University of Washington. However, these embeddings did not result in better performance, considered that they are smaller data quantity compared to the official GloVe embeddings.
As for the prosodic features, they used word time information described by the interval between the start of the current word and start of the previous word, and duration of the current word.
The models are trained on the Fisher English corpus \parencite{cieri-etal-2004-fisher}, which includes punctuation annotation and capitalization. The training dataset consists of 348 hours of conversational speech and the dev and test sets each consists of around 42 hours.
Only blanks (no punctuation), periods, commas and question marks were evaluated as the other punctuation classes were converted to blanks due to their low frequency. The dataset has a heavy class imbalance (Blank - 79.1\%; Comma - 11.5\%; Period - 8.2\%; Question Mark - 1.2\%).
They trained four models and present the results of per-class precision, recall and F1-score achieved by the CNN and Bi-LSTM models with and without time features to evaluate the contribution of prosodic features.
The best model in terms of F1 is the Bi-LSTMs with prosodic features: Comma achieves  66.1\% F1; Period  achieves  67.3\% F1 and Question Mark achieves  59.2\% F1. However, the punctuation predicted by the CNN with prosodic features is more accurate (better precision values), especially in the case of question marks in which this model achieves 72.9\% of Precision.

\textcite{sunkara20_interspeech} identified problems with several multimodal works of the literature, e.g. the model training is still suboptimal due to lack of large-scale parallel audio/text corpora. In their work they propose a novel framework for multimodal fusion of lexical and acoustic embeddings for punctuation prediction in conversational speech called multimodal semisupervised learning architecture (MuSe). The architecture contains three main components: acoustic encoder, lexical encoder, and a fusion component responsible to combine outputs from both the encoders. While previous approaches used word-level prosodic inputs and concatenated with lexical inputs, they argue that this mechanism does not capture the acoustic context beyond a word. Therefore, their model uses frame-level acoustic features for fusion with sub-word lexical encoder using two different approaches:  force-aligned word durations and sub-word attention model.
Like \textcite{ZelaskoSMSCD18}, they also conducted experiments on  Fisher English  corpus. It is important to comment that the three types of acoustic features used in the experiments are very elaborated: wav2vec features (for the feature extraction they trained a wav2vec-large model\footnote{https://github.com/pytorch/fairseq/blob/master/examples/wav2vec} on the 348-hour Fisher audio corpus) and two other prosodic features: pitch and melspec, computed using a 25ms frame window with 10ms frame shift. They extracted F0 features based on Kaldi pitch tracker method \parencite{pitch_2014} and used 80-dimension mel-scale spectrograms as alternative to pitch features. 
They compared the MuSe model with a Bi-LSTM based on the work of \textcite{ZelaskoSMSCD18} and they trained another lexical-only model which is a pretrained truncated BERT model. 
Pure lexical BERT model outperformed Bi-LSTM in all punctuation marks; when comparing lexical-only models with MuSe the results show that using any kind of acoustic feature (pitch, melspec or wav2vec) improves prediction of the all three classes. Comparing acoustic features, pitch and melspec have shown similar performance improvements, except for Question marks when attention is used for fusion. Unsupervised wav2vec features proved to be the best feature among all acoustic features for multimodal fusion. Comparing fusion techniques, forced-alignment  fusion (FA) performs slightly better than attention-based fusion (Att) for Fullstop while the performance is similar on Comma and Question Mark. Although \textcite{sunkara20_interspeech}  results are not directly comparable with those of \textcite{ZelaskoSMSCD18} on Fisher corpus  as the splits are different,  \textcite{sunkara20_interspeech}  achieved better performance in all classes of punctuation. The best individual F1 scores for Comma,  FullStop and  Question Marks are: 75.6\%, 75.6\% and 81.3\%, respectively.

Regarding the use of a cascaded approach, there are some problems in using the ASR output into a punctuation model based on a pre-trained model, such as BERT or some other transformer based model, that needs tokenization according to the vocabulary of the pre-trained model. Tokenizing texts without punctuation  and with possible ASR errors can cause some tokenization errors. To solve this problem \textcite{nozaki22_interspeech} proposed an end-to-end model for speech-to-punctuated-text recognition. They used stacked Transformer Encoder layers as a model architecture
and train it with a CTC loss function \parencite{graves_2006}
using speech as input and punctuated texts as output. For punctuation marks in English, they consider commas, periods, and question marks. They used two datasets of different languages: the multilingual MuST-C corpus\footnote{MuST-C comprises audio recordings from English TED Talks aligned at the sentence level with their manual transcriptions and translations.} \parencite{di_gangi_etal_2019_must} was used as the English dataset and the JCALL, an in-house Japanese dataset, consisting of audio recordings of conversations.
The evaluation metric used for punctuation prediction was the F1 score for each type of punctuation mark and its average were calculated. They first aligned the predicted text with the ground-truth text, then calculated the F1 score, in order to overcome ASR errors. Their model  can utilize acoustic information for punctuation prediction in order to be robust against ASR errors. The result of macro F1 score\footnote{Macro F1 gives each class equal importance, calculating the mean of the classes.} (for English) (average of the comma, period and question marks results) is 75.9\% and individual results using F1-score are: 61.1\% for commas, 92.5\% for period and  74.1\% for question marks. However,  this results are not comparable to the previous works for being calculated on different datasets.

OpenAI company launched in September of 2022 an open-source system for automated transcription. While it is also an end-to-end model, similar to the approach by \textcite{nozaki22_interspeech}, it has two important differences: it is open-source and was trained on large and multilingual data.  It is  an ASR capable of including punctuation and capitalization in the transcription \parencite{radford2022robust}.

Tables \ref{tab:table_211} and \ref{tab:table_212} present a summary of the revised works on punctuation prediction, highlighting the language evaluated in the study, domains (prepared or spontaneous speech), punctuation marks, evaluation datasets, metrics used, and the results, although they are not comparable among the works. 

\begin{table}[htpb]
\footnotesize
\centering
\caption{Summary of punctuation prediction works on prepared and spontaneous speech. Recent works on punctuation prediction usually rely on the four labels (Comma, Period, Question, O -- any other token) of the datasets proposed by the International Workshop on Spoken Language Translation 2011 (IWSLT 2011).}
\label{tab:table_211}
\begin{tabular}{l|l|l|l}
Source                      & Language(s) & Domain & Marks  \\
Alam, Khan, and Alam (2020) & English, Bangla  &   prep. speech      &   , . ?  O        \\
Yi and Tao (2019)           &    English         &  prep. speech       &  , . ?   O       \\
Zelasko et al. (2018)       &    English          &   spont. speech      &  , . ?   O    \\
Sunkara et al. (2020)       &      English        &   spont. speech      &   , . ?   O     \\
Nozaki et al. (2022)        &   English, Japanese          &  prep. speech       & , . ?   O       \\
Our Work on Whisper         &     Portuguese        &   spont. speech      &  , ... ! . ? ; :     \\
\end{tabular}
\end{table}

\begin{table}[htpb]
\footnotesize
\centering
\caption{This Table is a continuation of the Table \ref{tab:table_211} and presents the results in terms of F1 score for each punctuation mark common to all papers (comma, period, question mark). Size (in words) and results on IWSLT 2011 test sets are for manual (Reference) and ASR transcriptions, in this order. }
\label{tab:table_212}
\begin{tabular}{l|l|l|l|l}
 Test Set      & Duration/Size   & F1 (Comma)             &  F1 (Period)            & F1 (Question)\\
 IWSLT 2011    & 12,626/12,822 & 76.7\%; 66.3\%  &  88.9\%/82.3\%   & 87.8\%/63.4\%\\
 IWSLT 2011    & 12,626/12,822 &  64.1\%; 61.7\% &  79.9\%/75.6\%    &  74.8\%/69.1\%  \\
 Fisher Corpus &  1,100 utt.         &  66.1\%          &  67.3\%            & 59.2\% \\
 Fisher Corpus &   42 h          &  75.6\%          &  75.6\%            &  81.3\%\\
  MuST-C       &   2,641 utt.        &    61.1\%         &  92.5\%            &  74.1\% \\
  MuPe test set  &  16:48:34 h                   & 77.3\%          &  62.3\%            &  67.6\% \\
\end{tabular}
\end{table}

\subsection{Transcript-based Video Topic Modeling}
\label{sec:sec22}

Topic modeling is a popular method for organizing large volumes of data into a reduced set of subjects and themes \parencite{vayansky2020review}. Although it is more popular for textual data, in recent years, topic modeling has been applied to various tasks, such as audio and video applications \parencite{basu2016fuzzy, bleoancua2020lsi, thies2021graphtmt}. Topics represent high-level themes automatically discovered from correlations and similarities from low-level features. For example, we can extract topics from a video with a long interview about an artist's life story, such as childhood, education, family, and profession. Especially for video and audio documents, topic modeling is potentially useful to support summarization tasks, in which we can select representative audio and video segments from each topic to obtain a more concise final document \parencite{panda2017sparse}.

We can use different modalities of a video for topic modeling, such as visual, acoustical, and textual modalities. We focus on topic modeling that exploits textual information, especially on transcripts that are automatically extracted from audio. Topic modeling considering transcripts is not a recent idea. \textcite{taskiran2001automated} discuss the role of various modalities in organizing videos into themes and show that transcription achieves promising results, especially when combined with techniques to detect sentence boundaries. This strategy became more popular with the advent of video streaming platforms such as YouTube. \textcite{morchid2013lda} applied topic modeling to transcripts automatically generated by Youtube. The authors discuss the strengths of this strategy to identify tags in videos but emphasize the low performance of the model due to errors in the automatic transcription process. To deal with this challenge, \textcite{basu2016fuzzy} improve transcription data using spell checks and stopword removal, thereby showing the importance of transcription quality for topic modeling. 

An important aspect that impacts the performance of tasks involving video transcripts is the lack of textual punctuation \parencite{survey_2022}. Automatic transcription generation systems proposed in the last decade do not provide correct punctuation; therefore, there is an absence of fundamental text structures, such as sentences and phrases. This aspect negatively impacts the performance of topic modeling based on pre-trained language models, such as BERT, which depend on context and text structure. Although there are some proposed alternatives to mitigate this problem, as methods to infer periods, commas, questions, and exclamations in transcripts generated by automatic speech recognition systems, such methods are language dependent and contain parameters that are difficult to define in the training stage.

This paper focuses on the Whisper model for video transcription since it is a recent automatic speech recognizer that already incorporates punctuation identification for multiple languages. In this way, we can take advantage of Whisper sentence structures to identify well-formed audio segments, avoiding truncations in the video obtained after selecting representative segments of each topic. 

\section{OpenAI's Whisper} 
\label{sec:sec3}

Recently the ASR systems achieved great progress with the emergence of unsupervised pre-training techniques making it possible to pre-train models using over 1.000.000 hours of speech in different languages. An example of this technique is used on the model Wav2vec 2.0 \parencite{baevski2020wav2vec}. During the pre-training phase, the Wav2vec 2.0 model is trained using raw audio and it does not need the transcriptions. This is interesting because it is easy to find audio without transcriptions on the Internet. After this phase, the model is fine-tuned in smaller datasets that have transcriptions in this way learning the ASR task.

Although these pre-trained audio techniques make it possible to learn high-quality speech representations, they do not enable the neural model to learn a mapping of these representations to usable outputs (text) of equivalent performance, necessitating the fine-tuning in a smaller labeled dataset to actually perform the ASR task.
Unfortunately, this limits its usefulness and impact. Machine learning methods are prone to finding patterns in a dataset that increase performance on data retained from the same dataset. However, some of these patterns are brittle and spurious and do not generalize to other datasets and distributions. During the fine-tuning phase, the model can learn  dataset-specific quirks and although it achieves very good performance on the training dataset it can still make many basic errors when evaluated on another dataset with different recordings conditions and domains \parencite{geirhos2020shortcut}.

\subsection{Whisper Features}
\label{sec:sec31}

To alleviate this issue, researchers proposed the Whisper \parencite{radford2022robust}. Whisper  is a deep learning-based model for robust speech recognition. Whisper was trained with a multilingual dataset composed of 680.000 hours of labeled audio data. This dataset includes 563.000 hours of English speech and 117.000 hours covering 96 other languages. For Portuguese, it includes approximately 9 thousand hours of speech. The dataset was built using audio that is paired with transcripts on the Internet. In this way, the dataset is very diverse covering a broad distribution of audio from many different environments, speakers,  recording setups, and languages. In general, most ASR systems are trained with normalized text, where punctuation is removed and the transcription is all-uppercase or all-lowercase. Despite this, the Whisper model was trained with unnormalized transcriptions and it can predict automatically the transcription's punctuation.

Whisper was trained to transcribe and translate audio. The authors created a token that controls what task the model will perform. In addition, they added a token to indicate when has no speech in  the audio. A third token was used to control the language in  multilingual training. In this way, the Whisper can be used for transcribing, translating, language identification, and voice activity detection tasks. Finally, different from most ASR systems, Whisper can predict punctuation precisely.

There are nine models of different sizes (tiny, base, small, medium, large)  and approaches used in the training (using only English data and multilingual data). In our experiments we used the large model, with multilingual data.

\subsection{Illustrating the use of Whisper for Punctuation Prediction in Portuguese }
\label{sec:sec32}

The following excerpt from a MuPe life history\footnote{museudapessoa.org/historia-detalhe/?id=7853} presents four turns of an interview in which the punctuation marks were removed and the first words of each sentence are presented in lowercase.  
This excerpt helps to emphasize that the absence of punctuation can make it difficult for people to understand the text, since spontaneous speech can be segmented in different ways, as it does not follow the rules of written texts. This absence also can affect the task of modeling topics from audios, evaluated in this chapter. Generating video snippets for a topic produces a garbled result lacking coherence when no punctuation is available. 

\vspace{0.5cm}
\begin{description}
\item[Question]  qual é a origem da sua família ela é de São Paulo mesmo
\item[Answer] é é de São Paulo é quatrocentona (RISO) e é de São Paulo e lá acho que do tempo de dos bandeirantes sei lá eu não sei essa história muito antiga lá do tempo dos bandeirantes eu não sei e veio de Portugal teve um português lá que veio pro Brasil ficou aqui em São Paulo e daqui a família
\item[Question] seus avós também são de São Carlos
\item[Answer] não são de São Paulo assim meu pai nasceu em Itatiba minha mãe assim por acaso em Jacutinga Minas e que ela não aceitava de jeito nenhum porque meu avô tinha ido passar fazer um trabalho lá foi com a minha avó ela nasceu e ela não admitia isso ela era paulista então só eu da minha família só eu era meus pais tinha quatro irmãos nascidos aqui aí na revolução de 30 meu pai perdeu o emprego aqui e escolheu morar em São Carlos aí foi que eu nasci lá bem bastante diferença de idade na época a minha irmã a caçula da época tinha sete anos quando eu nasci então
\end{description}

The following output was generated by Whisper for the audio related to the same  MuPe life story shown above. 

\vspace{0.5cm}
\begin{description}
\item[Question] Qual é a origem da sua família?
 Ela é de São Paulo mesmo?
\item[Answer] É, é de São Paulo, é de 400 anos.
É de São Paulo e ela acho que de tempo dos Bandeirantes, sei lá, tem um aqui.
Eu não sei essa história muito antiga lá do tempo dos Bandeirantes, eu não sei.
E veio de Portugal, teve um português lá que veio para o Brasil, ficou aqui em São Paulo e daqui a família.
\item[Question] Seus avós também são de São Carlos?
\item[Answer] Não, são de São Paulo assim, meu pai nasceu em Itatiba. Minha mãe assim, por acaso nasceu em Jacutinga, Minas, e que ela não aceitava de jeito nenhum, porque meu avô tinha ido fazer um trabalho lá, foi com a minha avó e lá ela nasceu. E ela não admitia isso, ela era paulista.
Então, só eu, da minha família só eu.
Era meus pais, tinha quatro irmãos nascidos aqui, aí na Revolução de 30, meu pai perdeu o emprego aqui e escolheu morar em São Carlos. Aí foi que eu nasci lá, bastante diferença de idade. Na época, minha irmã, a caçula da época, tinha sete anos quando eu nasci.
\end{description}

When Whisper output is compared with the punctuated and capitalized manual transcription (shown below) we note that Whisper generates shorter sentences and thus more sentences (14) versus (7) than the manual transcription. However, the distribution of the punctuation used is very similar to the manual one (see Table \ref{tab:table_example}).  As for capitalization, it is mainly used for named entities related to cities, countries, states, regions. 

\vspace{0.5cm}
\begin{description}
\item[Question]  Qual é a origem da sua família, ela é de São Paulo mesmo?
\item[Answer]  É, é de São Paulo, é quatrocentona (RISO), e é de São Paulo e lá acho que do tempo de, dos bandeirantes, sei lá, eu não sei essa história muito antiga, lá do tempo dos bandeirantes, eu não sei e veio de Portugal. Teve um português lá que veio pro Brasil, ficou aqui em São Paulo e daqui a família...
\item[Question] Seus avós também são de São Carlos?
\item[Answer]  Não, são de São Paulo, assim, meu pai nasceu em Itatiba, minha mãe, assim, por acaso em Jacutinga, Minas e que ela não aceitava de jeito nenhum, porque meu avô tinha ido passar, fazer um trabalho lá, foi com a minha avó, ela nasceu e ela não admitia isso, ela era paulista, então só eu, da minha família só eu. Era meus pais, tinha quatro irmãos nascidos aqui, aí, na revolução de 30 meu pai perdeu o emprego aqui e escolheu morar em São Carlos, aí, foi que eu nasci lá, bem, bastante diferença de idade. Na época a minha irmã, a caçula, da época tinha sete anos quando eu nasci, então...
\end{description}

\begin{table}[ht]
\footnotesize
\centering
\caption{Comparing Whisper output for a manual transcription for punctuation marks and capitalization.}
\label{tab:table_example}
\begin{tabular}{l|l||l}
                    & Whisper Output      & Manual Transcription    \\ \hline
              
Comma                  &    24              &     34             \\
Ellipsis               &     0             &         2         \\
FullStop               &      11            &           9       \\
Question               &         3         &            2      \\ \hline
First word of sentences &        14       &          7         \\
Location Named-Entities &        11        &          11        \\
Other capitalizations   &         3         &          0        \\
\end{tabular}
\end{table}

Although Whisper capitalized the word ``Bandeirantes'' (there are several types of named entities called ``Bandeirantes'', such as a highway, a radio station, a TV channel), in the manual transcription it was not.  In the context of the interview this word refers to explorers (\textit{sertanistas}) from the colonial period in Brazil.

\section{Experimental Setup} 
\label{sec:sec4}

The Whisper model used in all the experiments was the multilingual large (1550 M parameters) from the official codebase\footnote{https://github.com/openai/whisper/blob/main/model-card.md}. Whisper was called with the parameters: language=Portuguese, temperature=0.15;
no\_speech\_threshold=0.6; e
condition\_on\_previous\_text=True.

Following the literature, we use Precision, Recall and F1-score for each evaluated punctuation and also calculated F1-macro to present the overall performance of the five types of punctuation marks that Whisper is able to generate on MuPe dataset. 
Even though punctuation datasets are generally imbalanced, macro F1 score still reflect true model performance.

\subsection{Evaluation Dataset} 
\label{sec:sec41}

The evaluation dataset in our work is composed of 10 life narratives taken from the Ponto de Cultura project of the MuPe  platform. There are 280 life narratives in this project, most of them with content available (several including transcription and full video available) and revised on the web platform, which were digitized and edited as part of the project 25 years of the MuPe in Brazil: Strengthening and Consolidation of Assets (PRONAC 164380)\footnote{https://acervo.museudapessoa.org/pt/apoie/quem-apoia/apoio-bndes}.

The collection of the MuPe is made up of life narratives, told by the people themselves or by third parties. The narratives are recorded in three ways: (i) at the Museum's headquarters, in a studio — recorded on video and collected by interviewers specialized in life-history methodology, (ii) sent via the internet by \textit{Programa Conte sua História} or (iii) via \textit{Museu que Anda}, a program in which the narratives of people outside the headquarters are recorded through itinerant booths. Each interview constitutes a unit of the collection that is formed by the audio or video recording of the interview, the transcription and edition of each narrative, accompanied by photos and documents sent by the people who tell their life narratives.

Once recorded, life narratives collected by MuPe are transcribed and sometimes revised. The transcripts have annotations of laughter, clapping hands, whistles, emotional speech, pauses, among others, using parentheses. Also, expansions of acronyms are annotated using square brackets. Moreover, the transcription is segmented and a proposal of punctuation is done using seven punctuation marks (see Table \ref{tab:table_stats_Punc}). The turns are indicated by P/1 (and P/2) and R labels followed by the transcription of the turn, where P/i (i = 1 or 2) indicates the interviewer (1 or 2 interviewers) and R the interviewee. However, since disfluencies, corrections and repetitions that are common in spontaneous speech are not annotated, the MuPe transcrition can be called an adapted verbatim transcription.

Table \ref{tab:table_stats_MuPe} shows statistics of the dataset, divided in two samples: male and female narratives. MuPe dataset is composed of 1,349 turns and totals 16:48:34 hours.

We show five statistics for manual transcriptions and also for Whisper outputs, as they differ. The number of sentences generated by Whisper is 26.6\% greater than the number of the manually segmented dataset, and therefore  
the average sentence length of Whisper output is smaller (13 words, without counting punctuation). 
We consider as sentence the segments ending with question, exclamation and fullstop marks, i.e. they carry a complete idea.

\begin{table}[ht]
\footnotesize
\centering
\caption{Data Statistics of MuPe samples: manual transcriptions and Whisper outputs. Average turn length and average sentence length are calculated in words, without counting punctuation. We consider as sentence the segments ending with question, exclamation and fullstop marks. Number of tokens includes punctuation.}
\label{tab:table_stats_MuPe}
\begin{tabular}{l|l||l|l}
                        & Male Sample          & Female Sample     &  Total       \\
Audio Duration          &    8:06:21 h         &   8:42:13         &   16:48:34 h \\ \hline
\multicolumn{4}{c}{Manual Transcription}                                               \\        \# Turns           &    834             &    515            &  1,349        \\ 
Average Turn Length   &  85.26 $\pm$ 169.44   &  138.80 $\pm$ 325.46 &  105.11 $\pm$  240.72 \\
\# Sentences            & 4,100               & 4,640             &       8,740     \\
Average Sentence Length  &   17.16 $\pm$ 27.47 &  14.57 $\pm $ 14.64  & 15.79  $\pm$ 21.67  \\
\#Tokens                &   83,953            &     79,377      &   163,330 \\ \hline
\multicolumn{4}{c}{Whisper Outputs}                                               \\  
\# Turns                &     931              &    598                  &  1,529        \\ 
Average Turn Length     &  77.40 $\pm$ 157.15  &  120.31 $\pm$ 294.86    &  94.18  $\pm$  222.35 \\
\# Sentences            & 5,522                 & 5,540                  &       11,062     \\
Average Sentence Length &   13.05 $\pm$ 11.52   &  12.98 $\pm $  11.46   & 13.02  $\pm$ 11.49 \\
\#Tokens                & 87,499             &  86,972                   &  174,471 \\
\end{tabular}
\end{table}

The number of tokens in Table \ref{tab:table_stats_MuPe} is substantially higher in Whisper than in manual transcription. Table \ref{tab:Tokens} shows that the difference mainly lies in the comma and fullstop punctuation tokens and in the common and high-frequency words of the language.

\begin{table}[ht]
\footnotesize
\centering
\caption{Descending count of tokens that show the biggest difference between Whisper's automatic transcription and manual transcription.}
\label{tab:Tokens}
\begin{tabular}{l|l||l|l}
\textbf{Token} & \textbf{Whisper} & \textbf{Manual Transcription} & \textbf{Difference} \\
,              & 18100        & 14954           & 3146          \\
.              & 10904        & 8350            & 2554          \\
para           & 1741         & 983             & 758           \\
a              & 4829         & 4331            & 498           \\
eu             & 4643         & 4279            & 364           \\
que            & 5254         & 4901            & 353           \\
então          & 1305         & 979             & 326           \\
o              & 3412         & 3134            & 278           \\
de             & 3894         & 3700            & 194           \\
aí             & 967          & 814             & 153           \\
não            & 2359         & 2208            & 151           \\
?              & 1392         & 1274            & 118           \\
ele            & 1119         & 1004            & 115           \\
porque         & 1041         & 927             & 114           \\
uma            & 1850         & 1738            & 112           \\
está           & 268          & 158             & 110           \\
era            & 1563         & 1459            & 104          \\
\end{tabular}
\end{table}

Table \ref{tab:table_stats_Punc}
 shows the distribution of seven punctuation classes in the manual transcription (MT) and in the Whisper output (WO).
The distribution of exclamation, semicolon and colon are higher in the manual transcriptions. As a matter of fact, Whisper is not able to generate semicolon and colon marks. The high numbers of question marks in both manual transcriptions and Whisper output is due to the fact that MuPe samples are made of interviews where MuPe interviewers ask several questions about the interviewee life story.

\begin{table}[ht]
\footnotesize
\centering
\caption{Distribution of punctuation classes in MuPe samples. Columns 2-4 present values of the reference dataset (manual transcription); columns 5-7 present values of Whisper ASR. Whisper is not able to generate semicolons and colons.  
Male Sample (MS); Female Sample (FS). }
\label{tab:table_stats_Punc}
\begin{tabular}{l|l|l|l||l|l|l}
  \multicolumn{4}{c}{Manual Transcription} & \multicolumn{3}{c}{Whisper Output} \\
               & MS   & FS   & Total & MS    & FS & Total    \\
               & \#      & \#      & \# (\%)  & \#     & \#   & \# (\%)\\  \hline                                          
Ellipsis       &  257   &     110     &  367 (1.56\%)  & 303   & 266   & 569 (1.90\%)\\ 
Exclamation    &    45  &     172     &  217 (0.92\%) & 31    & 34    & 65 (0.22\%) \\ 
FullStop       &  3,247 &    4,002    &  7,249 (30.8\%)    & 4,656 & 4,949 & 9,605 (32.2\%)\\ 
Question       &  808   &      466    & 1,274 (5.41\%)    & 835   & 557   & 1,392 (4.66\%)\\ 
Comma          & 8,383  &    6,571    &  14,954 (63.5\%)  & 9,346 & 8,879 & 18,225 (61.04\%)  \\ 
Semicolon      &  62    &     21      &   83 (0.35\%)      & 0     & 0     & 0 \\ 
Colon          &   293  &    358      &   651 (2.76\%)     &  0    & 0     & 0   \\ \hline
Total          &        &              &  23,521    &       &       &  29,856\\
\end{tabular}
\end{table}

\subsection{Data Preparation} 
\label{sec:sec42}
\subsubsection{Data Preparation for Punctuation Analysis} 
\label{sec:sec421}

For punctuation analysis, we processed the automatic transcription and aligned it with the original samples of MuPe, our reference dataset. We chose to diarize\footnote{We used the Pyannote \parencite{Bredin2020} tool to diarize the audios.} the original audios to further improve the quality of the alignment process as we can work with smaller segments of audio. After the diarization, we generate the automatic transcription using the Whisper ASR.

It is important to note that the diarization process can affect the output of Whisper. We noticed 12 differences in the excerpt of Section \ref{sec:sec32}, and also appears the word ``Então'' followed by ellipses in the end of the excerpt, like in the manual transcription.  The following output was generated by Whisper when applied to each segment generated by the diarization tool. We show, in bold, different punctuation marks in the diarized narrative compared to the Whisper output shown in Section \ref{sec:sec32}. We show the punctuation marks along with the word that precedes the punctuation for easy comparison.

\vspace{0.5cm}
\begin{description}
\item [Question] Qual é a origem da sua família?
Ela é de São Paulo \textbf{mesmo.}
\item [Answer] É de São Paulo, é de Quatrocentona.
É de São Paulo e é lá de tempo de... dos \textbf{Bandeirantes.}
Sei lá, tem um aqui.
Não sei essa história muito \textbf{antiga,} lá do tempo dos Bandeirantes, não sei.
E veio de \textbf{Portugal.}
Teve um português lá que veio para o Brasil, ficou aqui em São Paulo e daqui a família.
\item [Question] Seus avós também são de São Carlos?
\textbf{Não.} 
\item [Answer] São de \textbf{São Paulo,} meu pai nasceu em Itatiba. Minha \textbf{mãe,} por \textbf{acaso,} nasceu em Jacutinga, Minas, e ela não aceitava de jeito nenhum, porque meu avô tinha ido fazer um trabalho lá, foi com a minha avó e lá ela nasceu. E ela não admitia isso, ela era paulista. Então, só eu, da minha \textbf{família,} só eu. Meus pais, tinha quatro irmãos nascidos \textbf{aqui.} \textbf{Aí,} na Revolução de 1930, meu pai perdeu o emprego aqui e escolheu morar em São Carlos. Aí foi que eu nasci \textbf{lá.}
Bastante diferença de idade. Na época, a minha irmã, a caçula da época, tinha sete anos quando eu nasci. \textbf{Então...} 
\end{description}

As automatic transcriptions can contain errors, it can be difficult to perfectly align all the sentences. However, since the original audio is composed by several turns of an interview in which each speaker makes a question or answers it, the diarization process generates small parts of texts which facilitated the alignment. 

We removed all the quotation marks of the MuPe reference dataset, in a preprocessing step, to be fair with the automatic transcription as Whisper is not able to generate quotation marks to enclose direct speech, e.g.:   
\vspace{0.5cm}
\begin{description}
    \item[Direct Speech in MuPe] Liguei pra ele e disse: “olha, o que você acha? Na ECA tem gente muito boa, mas eu não queria fazer ligado ao palco, queria formação”.
    \item[Preprocessing] Liguei pra ele e disse: olha, o que você acha?
Na ECA tem gente muito boa, mas eu não queria fazer ligado ao palco, queria formação.
\end{description}

We also performed other preprocessing steps in both original and automatic transcription texts, creating a list of segments in which each segment ends with a punctuation\footnote{The list of punctuation used in the experiments are those seven presented in Table \ref{tab:table_stats_Punc}.}. We chose to do this preprocessing to improve the robustness of the aligner and also to facilitate the punctuation analysis, since all segments end with a punctuation, which can be used to compare whether the automatically and manually generated punctuation match. After the preprocessing step, we performed the alignment. We created a specific alignment tool for this task. Our algorithm is composed by three steps:

\begin{enumerate}
    \item \textbf{Match text}: we provide a list of automatic transcription segments and manual segments (a segment ends with one of the seven types of punctuation considered in this work). The algorithm calculates the difference between each segment and selects the best match using a scoring metric;
    \item \textbf{Preliminary alignment}: the algorithm iterates over the segments of the manual dataset, and for each segment, iterates over the segment of the matched automatic transcription segment of the previous step. If some manual segment matches the automatic transcription with some small degree of error, the algorithm align the two segments. 
    In the perfect scenario, each manual segment should match a segment of the automatic transcription.  However, in our scenario we observe that manual transcriptions have longer segments showing complete ideas --- called here sentences (see Table \ref{tab:table_stats_MuPe}) --- than automatic ones. In these cases, it is necessary to align the segments using another strategy;
    \item \textbf{Contextual alignment}: the remaining segments that missed alignments were aligned as follows: we selected the most similar segment from the automatic transcription segments and added context using their segment neighbors, that is, we added the segments before and after the current segment that is most similar to the segment being aligned. Then, we performed the removal of words in the start and the end of the segments until the best score is obtained. Here, we calculated the matching score using Levenhstein distance and Longest Common Subsequence (LCS).
\end{enumerate}

The alignment using the metrics Levenshtein and LCS was very similar. However, here we chose to use both of them by averaging the two scores. We normalize the segments during the alignment process (lowercase the text, remove the punctuation and convert numbers to strings) to achieve best results, while maintaining the original transcription of MuPe to recover the punctuation after the alignment process. On average, 85,44\% of MuPe segments were aligned.

With the aligned segments we were able to measure Word Error Rate  (WER) and Character Error Rate (CER) values for the Whisper model on MuPe test set. Whisper model achieved a WER of 14.50\% on MuPe test set. We also use the metric CER, because for smaller audios, with just a few words, this metric tends to be more reliable. When measuring CER, it was obtained 8.13\% on MuPe test set. Regarding the male sample, the values for WER and CER are 12.97\% and 7.37\%, respectively; and for the female sample, the values for WER and CER are approximately 3\% and 1.5\% worse than the values for the male sample (16.14\% and 8.95\%, respectively).

\subsubsection{Data Preparation for Transcript-based Video Topic Modeling} 
\label{sec:sec422}

We use the automated transcripts generated by Whisper for transcript-based video topic modeling. The generated punctuation marks are used to define video segments. Thus, each video $V$ is defined as a set of $k$ segments, $V = (\vec{s}_1,\vec{s}_2,...,\vec{s}_k)$. The textual information of the segments is used to obtain a vector-space model representation through a language model based on BERT, more specifically, the \textit{paraphrase-multilingual-mpnet-base-v2} model available in the SBERT project \parencite{reimers-2019-sentence-bert}. Following recent topic-modeling strategies based on language models, we assume similar segments can be allocated to the same topic \parencite{sia2020tired}.

A topic is defined as a set of related segments, i.e., neighbors in the vector space according to some measure of similarity or distance. We use supervised topic modeling, in which we have reference topics extracted from the Media Topics Taxonomy of the International Press Telecommunications Council (IPTC)\footnote{https://iptc.org/standards/media-topics/}. In this taxonomy, there are approximately $1200$ topics about different events and subjects. Each reference topic is also mapped onto the same vector space obtained by the language model. Thus, each video segment is associated with the most similar reference topic, according to the cosine similarity $cos(\vec{s}_i,\vec{t}_j) = {\vec{s}_i \cdot \vec{t}_j \over \|\vec{s}_i\| \|\vec{t}_j\|}$, where $\vec{s}_i$ is a segment of a video $V$ and $\vec{ t}_j$ is the $j$th reference topic.

The previous steps represented the video through segments and associated each segment with a reference topic according to textual similarity. Although these steps are potentially helpful in improving information retrieval systems from cross-topic videos, in this work we are interested in obtaining a summarized video of a MuPe life story interview. Thus, data preparation's last step is automatically selecting representative segments for each topic. In this case, we use a simple and intuitive strategy: select segments with the greatest cosine similarity to their topics until the output video reaches a size (in seconds) predefined by the user.

Note that the correct identification of punctuation marks generated by Whisper during ASR is essential for the entire transcript-based video topic modeling process. Errors in this step impair the association of segments with their topics and the quality of the summarized video with non-coherent and truncated segments.

\section{Results and Discussion} 
\label{sec:sec5}

\subsection{Results for Punctuation Prediction} 
\label{sec:sec51}

Our best results,
in terms of the overall F1 score, were for the comma punctuation (77.5\%), which has the higher distribution in the MuPe testset (see Table 5), and question mark (68\%), as shown in Table 6.

\begin{table}[]
\centering
\caption{General results. Whisper is not able to generate colons and semicolons. We leave these lines blank.}
\begin{tabular}{ll|l|l|l}
         &             & P (\%) & R (\%) & F1 (\%) \\ \cline{2-5} 
Pausing  & Comma       & 78.1 & 77.0    & 77.5     \\ \cline{2-5} 
Points    & Semicolon  & ---  & ---     &    ---          \\ \cline{2-5} 
          & Colon       &  --- & ---     &    ---         \\ \cline{2-5} \hline \hline
Complete & Exclamation & 17.2  & 3.0    & 5.1     \\ \cline{2-5} 
Ideas     & Question    & 71.5  & 64.9  & 68.0     \\ \cline{2-5} 
         & FullStop    & 59.4    & 69.2  & 63.9     \\ 
\cline{2-5} \hline \hline
         & Ellipsis     & 16.5    & 18.3    & 17.4    \\ \hline 
Average (\%)&          & 48.5 $\pm$ 29.7 &  46.5 $\pm$ 33.4 &  46.4 $\pm$ 32.7    \\
\end{tabular}
\end{table}

Although our results are not directly comparable with the results provided 
on Fisher corpus (a spontaneous speech corpus like our MuPe samples) in Table 1 and 2, we achieved better performance in two classes of punctuation, in terms of F1 scores: comma and fullstop.

Our best results for the female sample (Table 7),
in terms of the overall F1 score, were for the comma punctuation (75\%), which has the higher distribution in this sample (see Table 5), and question mark (69.1\%). Regarding exclamation marks, it has a low F1 score, but a precision higher than its recall. On average, precision is higher for the female sample than for the male sample, while F1 score values are similar for both samples.

\begin{table}[ht]
\centering
\caption{Results for female sample. Whisper is not able to generate colons and semicolons. We leave these lines blank.}
\begin{tabular}{ll|l|l|l}
         &              & P (\%)      & R (\%) & F1  (\%) \\ \cline{2-5}
Pausing  & Comma        & 73.5 & 76.5 & 75.0 \\ \cline{2-5}
Points    & Semicolon   &   ---&   ---&   ---        \\ \cline{2-5}
         & Colon        &   ---&  --- &      ---        \\ \cline{2-5}\hline \hline
Complete & Exclamation & 31.2 & 0.4 & 6.8 \\ \cline{2-5}
Ideas     & Question   & 72.7 & 65.8 & 69.1 \\ \cline{2-5}
         & FullStop    & 63.4 & 67.8 & 65.5 \\ \cline{2-5}\hline \hline
         & Ellipsis    & 12.4 & 17.6 & 14.5  \\ \hline
Average (\%) &         & 50.6 $\pm$ 27.4 &   45.6 $\pm$ 34.2 & 46.2 $\pm$ 32.7   \\   
\end{tabular}
\end{table}

\begin{table}[ht]
\centering
\caption{Results for male sample. Whisper fails to correctly predict exclamation marks and it is not able to generate colons and semicolons.}
\begin{tabular}{ll|l|l|l}
        &              & P (\%)     & R (\%)     & F1  (\%)    \\ \cline{2-5} 
Pausing  & Comma       & 82.2   &     77.4  &  79.7      \\ \cline{2-5}
Points   & Semicolon   &   ---  &     ---  &    ---           \\ \cline{2-5}
         & Colon       &   ---  &      ---   &    ---         \\ \cline{2-5}\hline \hline
Complete & Exclamation &    0.0 &        0.0    &  0.0            \\ \cline{2-5}
Ideas     & Question    & 70.7         & 64.3     & 67.4 \\ \cline{2-5}
         & FullStop    & 55.2           & 70.9      & 62.0 \\ \cline{2-5}\hline \hline
         & Ellipsis    & 19.5           & 18.7      &  19.1     \\ \hline
Average (\%) &         & 45.5 $\pm$ 34.7 &   46.2 $\pm$ 34.6  &  45.6 $\pm$  34.2\\    
\end{tabular}
\end{table}

The best results for the male sample (Table 8),
in terms of the overall F1 score, was for the comma punctuation (79.7\%), which has the higher distribution in this sample (see Table 5).

Regarding the failing in predicting exclamation marks it is worth investigating, as f0 
rises, in the case of exclamation, both in female and male voices. However, the female is, in general, more acute compared to male voice and Whisper may be sensitive to this variation. In addition, in the male sample, exclamation marks have the lower distribution.

\subsection{Results for Transcript-based Video Topic Modeling} 
\label{sec:sec52}

In the previous section, we analyzed Whisper's ability to identify punctuations for transcripts extracted from male and female audio. Now, we focus the analysis on a transcript-based video topic modeling application, in which punctuation is crucial to the whole process.

Remember that the proposed process generates a summarized video containing each topic's segments. For the experimental evaluation, we used 600 seconds (5 minutes) as a parameter to guide the number of representative segments. Thus, we have the original transcript $T_{in}$ generated by Whisper and the transcript $T_{out}$ obtained after the topic modeling process. Our objective is to evaluate how much the transcription $T_{out}$ preserves the information available in $T_{in}$.

We use the BLANC measure \parencite{vasilyev2020fill}, popular for summarization tasks, which allows us to measure how well $T_{out}$ summarizes $T_{in}$ through a neural language model. Although it is an indirect evaluation of the punctuation in the transcript-based topic modeling process, our assumption is that if the punctuation was correctly identified, then representative segments were identified, and consequently, coherent topics were generated. The Masked Language Model is a basic task in neural language models, in which a portion of tokens is initially masked, and the objective of the model is to use context to predict such tokens. BLANC is defined as $\frac{N_{help} - N_{base}}{N_{total}}$, where $N_{help}$ is the number of successfully unmasked tokens when the $T_{out}$ is used by the model; $N_{base}$ is the number of successfully unmasked tokens when the $T_{out}$ is not used by the model; and $N_{total}$ is the total number of masked tokens. Formally, the BLANC measure ranges from -1 to 1. In practice, with default parameters proposed by the BLANC authors, the measure usually values range from $0.05$ to $0.30$, in which values close to zero indicate that $T_{out}$ is useless for predicting masked tokens.

\begin{table}[ht]\centering
\caption{Results (BLANC scores) of transcript-based video topic modeling considering two scenarios. Bold values indicate better performance.}\label{tab: }
\begin{tabular}{lr|rr}
\textbf{} &\textbf{Transcripts with Punctuation} &\textbf{Transcripts without Punctuation} \\\midrule
\multirow{5}{*}{Female samples} &\textbf{0.1723} &0.1628 \\
&\textbf{0.1271} &0.1090 \\
&0.1185 &\textbf{0.1294} \\
&\textbf{0.1140} &0.0855 \\
&\textbf{0.1356} &0.1223 \\
\hline
\multirow{5}{*}{Male samples} &\textbf{0.1795} &0.1406 \\
&\textbf{0.1229} &0.1213 \\
&\textbf{0.1191} &0.1006 \\
&\textbf{0.1232} &0.1004 \\
&\textbf{0.1594} &0.1508 \\

\end{tabular}
\end{table}

Table 9 shows two transcript-based video topic modeling evaluation scenarios. In the first scenario (Transcripts with Punctuation), we use $T_{out}$ obtained by the process described in Section \ref{sec:sec422}, with all punctuations. The results of the BLANC measure are satisfactory compared to other works that applied such an evaluation, indicating that the transcript-based topic modeling obtains a resulting video that preserves relevant information from the original video. In particular, this is a promising result for the application of MuPe involving video interviews of people's life stories, as it allows a concise view of the main topics of the interview. In the second scenario (Transcripts without Punctuation), we also analyzed the impact of the score for the BLANC measure itself, simulating videos generated with truncated and non-coherent segments. In this case, there is a reduction observed in 9 of the 10 videos, thereby reinforcing the relevance of correct punctuation identification for this application.

\section{Conclusions and Future Work} 
\label{sec:sec6}

In this chapter, we present an evaluation of Whisper ASR in relation to identifying punctuation in Portuguese transcripts. In addition to measuring the precision, recall, and F1 of the comma, exclamation, question, and fullstop of Whisper transcripts against human annotation, we also discuss the impact of punctuation on a MuPe application of transcript-based video topic modeling.

Although Whisper still needs improvements for ASR in Portuguese regarding punctuation, we observed a significant advance when compared to previous proposals in the literature. In particular, we highlight the practical advantages of a pre-trained multilingual model. On the other hand, the results obtained here opened up new questions and research directions for future work, as listed below:

\begin{enumerate}
\item Evaluate whether there is a differentiated performance regarding the automatic prediction of the Whisper marks between textual genres of the NURC-SP Minimal Corpus, which contains 21 surveys of the three genres of the NURC-SP: conversations (inquiries of the type D2 and DiD) and formal elocutions, such as classes and lectures (EF-type inquiries);
\item Evaluate Whisper's performance in segmenting datasets annotated with terminal and non-terminal segments of the NURC-SP Minimal Corpus. It is expected that the exclamation and question marks, and also fullStop marks fall on the terminal segments and comma and ellipsis fall on the non-terminal segments;
\item Evaluate the cause of the different performance between genders (men and women) in the MuPe dataset, with the help of metrics such as pitch and the set of ASR wav2vec metrics as used in the work of \textcite{sunkara20_interspeech}; and
\item Extend MuPe's video topic modeling application by also considering sentiment analysis of parts of the interview. In particular, we intend to employ multimodal sentiment analysis considering both the content of the transcripts and the acoustic characteristics associated with the valence and arousal of the audio.
\end{enumerate}

\begin{acknowledgement}
This work is part of a Technology Transfer Agreement among Museum of Person (MuPe), Instituto de Ciências Matemáticas e de Computação da Universidade de São Paulo (ICMC/USP) and Federal University of Goiás.
This work was carried out at the Center for Artificial Intelligence (C4AI-USP), with support by the São Paulo Research Foundation (FAPESP grant \#2019/07665-4) and by the IBM Corporation. We also thank the support of the Centro de Excelência em Inteligência Artificial (CEIA) funded by the Goiás State Foundation (FAPEG grant \#201910267000527). This project was also supported by the Ministry of Science, Technology and Innovation, with resources of Law No. 8.248, of October 23, 1991, within the scope of PPI-SOFTEX, coordinated by Softex and published Residence in TIC 13, DOU 01245.010222/2022-44.
\end{acknowledgement}
\printbibliography
\end{document}